%% file: paper.tex
\documentclass[11pt]{article}
\usepackage{acl2016}
\usepackage{times}
\usepackage{latexsym}

\aclfinalcopy

\usepackage{multirow}
\usepackage{url}

\usepackage[ruled,vlined,linesnumbered]{algorithm2e}

\usepackage{graphicx}
\usepackage{amsmath,amsfonts,amsthm,amssymb}
\usepackage{dsfont}
\usepackage{booktabs}

\usepackage{mdwlist}
\usepackage{paralist}

\usepackage[font=small]{caption}
\input{dfn}

\newcommand{\embed}[1]{\textsc{e}(#1)}

\newcommand{\name}{\textsc{CFO}\xspace}
\newcommand{\acc}{75.7\%\xspace}
\newcommand{\improve}{11.8\%\xspace}
\newcommand{\sq}{\textit{single-fact questions}\xspace}

\makeatletter
\let\inserttitle\@title
\makeatother

\title{\name: Conditional Focused Neural Question Answering \\with Large-scale Knowledge Bases}

\author{
Zihang Dai\thanks{~~Part of the work was done while at Baidu.} \\
Carnegie Mellon University \\
\texttt{dzihang@andrew.cmu.edu}
\And
Lei Li$^*$ \\
Toutiao.com \\
\texttt{lileicc@gmail.com}
\And
Wei Xu \\
Baidu Research\\
\texttt{xuwei06@baidu.com}
}

\date{}

\begin{document}

\maketitle

\begin{abstract}
\input{000abstract}

\end{abstract}

\section{Introduction}
\label{sec:intro}
\input{010intro}

\section{Related Work}
\label{sec:related}
\input{020related}

\section{Overview}
\label{sec:overview}
\input{030overview}

\section{Proposed \name}
\label{sec:model}
\input{040model}

\subsection{Review: Gated Recurrent Units}
\label{sec:review_gru}
\input{041review_gru}

\subsection{Model Parameterization}
\label{sec:model_parameterization}
\input{042model_parameterization}

\subsection{Focused Pruning}
\label{sec:subject_labeling}
\input{043subject_labeling}


\section{Parameter Estimation}
\label{sec:training}
\input{050training}

\subsection{Decomposable Log-Likelihood}
\label{sec:decompose_likelihood}
\input{051decompose_likelihood}

\subsection{Approximation with Negative Samples}
\label{sec:hinge_loss}
\input{052hinge_loss}

\section{Experiments}
\label{sec:experiments}
\input{070experiments}

\subsection{Dataset and Knowledge Base}
\label{sec:dataset}
\input{071dataset}

\subsection{Evaluation and Baselines}
\label{sec:evaluation}
\input{072evaluation}

\subsection{Experiment Setting}
\label{sec:experiment_setting}
\input{073experiment_setting}

\subsection{Results}
\label{sec:results}
\input{074results}

\subsection{Effectiveness of Pruning}
\label{sec:eval_subject_labeling}

\input{075eval_subject_labeling}

\subsection{Additional Analysis}
\label{sec:additional_analysis}
\input{076additional_analysis}

\section{Conclusion}
\label{sec:conclusion}
\input{080conclusion}


{\small 
	\bibliography{BIB/ref}
	\bibliographystyle{acl2016}
}

\end{document}

%% file: dfn.tex
\newcommand{\ben}{\begin{enumerate*}}
\newcommand{\een}{\end{enumerate*}}
\newcommand{\beq}{\begin{equation}}
\newcommand{\eeq}{\end{equation}}
\newcommand{\bit}{\begin{itemize*}}
\newcommand{\eit}{\end{itemize*}}

\newcommand{\hide}[1]{}

\newcommand{\indep}[1]{\perp}

\DeclareMathOperator*{\argmax}{\arg\,\max}



%% file: 000abstract.tex
How can we enable computers to automatically answer questions like ``\textit{Who created the character Harry Potter}''?
Carefully built knowledge bases provide rich sources of facts. 
However, it remains a challenge to answer factoid questions raised in natural language 
due to numerous expressions of one question. 
In particular, we focus on the most common questions --- ones that can be answered with a single fact in the knowledge base.
We propose \name, a \textbf{C}onditional \textbf{Fo}cused neural-network-based approach to answering factoid questions with knowledge bases. 
Our approach first zooms in a question to find more probable candidate subject mentions, 
and infers the final answers with a unified conditional probabilistic framework. 
Powered by deep recurrent neural networks and neural embeddings,
our proposed \name achieves an accuracy of \acc on a dataset of 108k questions -- the largest public one to date. 
It outperforms the current state of the art by an absolute margin of \improve.

%% file: 010intro.tex
Community-driven question answering (QA) websites such as Quora,
Yahoo-Answers,
and Answers.com
are accumulating millions of users and hundreds of millions of questions. 
A large portion of the questions are about facts or trivia.
It has been a long pursuit to enable machines to answer such questions automatically.

In recent years, several efforts have been made on utilizing open-domain knowledge bases to answer factoid questions.
A knowledge base (KB) consists of structured representation of facts in the form of subject-relation-object triples. 
Lately, several large-scale general-purpose KBs have been constructed, including YAGO~\cite{suchanek2007yago}, Freebase~\cite{bollacker2008freebase}, 
NELL~\cite{carlson2010toward}, and DBpedia~\cite{lehmann2014dbpedia}. 
Typically, structured queries with predefined semantics (e.g. SPARQL) can be issued to retrieve specified facts from such KBs.
Thus, answering factoid questions will be straightforward once they are converted into the corresponding structured form.
However, due to complexity of language, converting natural language questions to structure forms remains an open challenge.


Among all sorts of questions, there is one category that only requires a single fact (triple) in KB as the supporting evidence.
As a typical example, the question ``\textit{Who created the character Harry Potter}'' can be answered with the single fact \texttt{\small(HarryPotter, CharacterCreatedBy, J.K.Rowling)}. 
In this work, we refer to such questions as \sq. 
Previously, it has been observed that \sq constitute the majority of factoid questions in community QA sites~\cite{fader2013paraphrase}.
Despite the simplicity, automatically answering such questions remains far from solved --- the latest best result on a dataset of 108k single-fact questions is only 63.9\% in terms of accuracy \cite{bordes2015large}.

To find the answer to a single-fact question, it suffices to identify the subject entity and relation (implicitly) mentioned by the question, 
and then forms a corresponding structured query.
The problem can be formulated into a probabilistic form.
Given a single-fact question $q$, finding the subject-relation pair $\hat{s}, \hat{r}$ from the KB $\mathcal{K}$ which maximizes the conditional probability $p(s,r|q)$, i.e.
\begin{equation}
\hat{s}, \hat{r} = \argmax_{s, r \in\mathcal{K}} p(s, r | q)
\label{eq:single-fact-qa-defn}
\end{equation}
Based on the formulation \eqref{eq:single-fact-qa-defn}, the central problem is to estimate the conditional distribution $p(s, r | q)$.
It is very challenging because of
\begin{inparaenum}[\itshape a\upshape)]
	\item the vast amount of facts --- a large-scale KB such as Freebase contains billions of triples,
	\item the huge variety of language --- there are multiple aliases for an entity, and numerous ways to compose a question,
	\item the severe sparsity of supervision --- most combinations of $s, r, q$ are not expressed in training data.
\end{inparaenum}
Faced with these challenges, existing methods have exploited to 
incorporate prior knowledge into semantic parsers,
to design models and representations with better generalization property, 
to utilize large-margin ranking objective to estimate the model parameters,
and  to prune the search space during inference. 
Noticeably, models based on neural networks and distributed representations have largely contributed to the recent progress (see section \ref{sec:related}).

In this paper, we propose \name, a novel method to answer \sq with large-scale knowledge bases.
The contributions of this paper are,
\begin{itemize*}
	\item we employ a fully probabilistic treatment of the problem with a novel conditional parameterization using neural networks,
	\item we propose the focused pruning method to reduce the search space during inference, and
	\item we investigate two variations to improve the generalization of representations for millions of entities under highly sparse supervision. 
\end{itemize*}
In experiments, \name achieves \acc in terms of top-1 accuracy on the largest dataset to date, outperforming the current best record by an absolute margin of \improve.


%% file: 020related.tex
The research of KB supported QA has evolved from earlier domain-specific QA \cite{zelle1996learning,tang2001using,liang2013learning} to open-domain QA based on large-scale KBs. 
An important line of research has been trying to tackle the problem by semantic parsing, which directly parses natural language questions into structured queries~\cite{liang11learning,cai2013large,kwiatkowski2013scaling,yao2014information}.
Recent progresses include 
designing KB specific logical representation and parsing grammar~\cite{berant2013semantic}, 
using distant supervision~\cite{berant2013semantic}, 
utilizing paraphrase information~\cite{fader2013paraphrase,berant2014semantic}, 
requiring little question-answer pairs~\cite{reddy2014large},
and exploiting ideas from agenda-based parsing~\cite{berant2015imitation}.

In contrast, another line of research tackles the problem by deep learning powered similarity matching.
The core idea is to learn semantic representations of both the question and the knowledge from observed data, such that the correct supporting evidence will be the nearest neighbor of the question in the learned vector space.
Thus, a main difference among several approaches lies in the neural networks proposed to represent questions and KB elements.
While \cite{bordes2014open,bordes2014question,bordes2015large,yang2014joint}
use relatively shallow embedding models to represent the question and knowledge, \cite{yih2014semantic,yih2015semantic} employ a convolutional neural network (CNN) to produce the representation. 
In the latter case, both the question and the relation are treated as a sequence of letter-tri-gram patterns, and fed into two parameter shared CNNs to get their embeddings.
What's more, instead of measuring the similarity between a question and an evidence triple with a single model as in \cite{bordes2015large}, \cite{yih2014semantic,yih2015semantic} adopt a multi-stage approach. 
In each stage, one element of the triple is compared with the question to produce a partial similarity score by a dedicated model. 
Then, these partial scores are combined to generate the overall measurement. 

Our proposed method is closely related to the second line of research, since neural models are employed to learn semantic representations.
As in \cite{bordes2015large,yih2014semantic}, we focus on \sq.
However, we propose to use recurrent neural networks (RNN) to produce the question representation.
More importantly, our method follows a probabilistic formulation, and our parameterization relies on factors other than similarity measurement.

Besides KB-based QA, our work is also loosely related to work using deep learning systems in QA tasks with free text evidences.
For example, \cite{iyyer2014neural} focuses questions from the quiz bowl competition with recursive neural network.
New architectures including memory networks~\cite{weston2015memory}, dynamic memory networks~\cite{kumar2015ask}, and more~\cite{peng2015towards,lee2015reasoning} have been explored under the \texttt{bAbI} synthetic QA task~\cite{weston2016towards}.
In addition, \cite{hermann2015teaching} seeks to answer Cloze style questions based on news articles.

%% file: 030overview.tex
In this section, we formally formulate the problem of {\em single-fact question answering} with knowledge bases.
A knowledge base $\mathcal{K}$ contains three components: a set of entities $\mathcal{E}$, a set of relations $\mathcal{R}$, and a set of facts 
$\mathcal{F} = \{\langle s, r, o\rangle\} \subseteq \mathcal{E}\times \mathcal{R}\times \mathcal{E}$, 
where $s, o \in \mathcal{E}$ are the subject and object entities, and $r\in \mathcal{R}$ is a binary relation. 
$\embed{r}, \embed{s}$ are the vector representations of a relation and an entity, respectively. 
$s \to r$ indicates that there exists some entity $o$ such that $\langle s, r, o \rangle \in \mathcal{F}$.
For single-fact questions, a common assumption is that the answer entity $o$ and some triple $\langle s_i, r_k, o \rangle \in \mathcal{F}$ reside in the given knowledge base.
The goal of our model is to find such subject $s_i$ and relation $r_k$ mentioned or implied in the question. Once found, a structured query (e.g. in SPARQL) can be constructed to retrieve the result entity.

\subsection{Conditional Factoid Factorization}
Given a question $q$, the joint conditional probability of subject-relation pairs $p(s, r| q)$ can be used to retrieve the answer using the exact inference defined by Eq.~\eqref{eq:single-fact-qa-defn}. 
However, since there can be millions of entities and thousands of relations in a knowledge base, it is less effective to model $p(s,r|q)$ directly. 
Instead, we propose a conditional factoid factorization,
\begin{equation}\label{eq:conditional-view}
p(s,r|q) = p(r|q) \cdot p(s|q,r)
\end{equation}
and utilize two neural networks to parameterize each component, $p(r|q)$ and $p(s|q, r)$, respectively. 
Hence, our proposed method contains two phases: inferring the implied relation $r$ from the question $q$, and inferring the mentioned subject entity $s$ given the relation $r$ and the question $q$. 

There is an alternative factorization $p(s, r| q) = p(s | q) \cdot p(r | s, q)$. 
However, it is rather challenging to estimate $p(s | q)$ directly due to the vast amount of entities ($>10^6$) in a KB. 
In comparison, our proposed factorization takes advantage of the relatively limited number of relations (on the order of thousands). 
What's more, by exploiting additional information from the candidate relation $r$, it's more feasible to model $p(s | q, r)$ than $p(s | q)$, leading to more robust estimation.

A key difference from prior multi-step approach is that our method do not assume any independence between the target subject and relation given a question, 
as does in the prior method~\cite{yih2014semantic}. 
It proves effective in our experiments. 

\subsection{Inference via Focused Pruning}
\label{sec:pruning-formulation}
As defined by the Eq.~\eqref{eq:single-fact-qa-defn}, a solution needs to consider all available subject-relation pairs in the KB as candidates. 
With a large-scale KB, the number of candidates can be notoriously large, resulting in a extremely noisy candidate pool.
We propose a method to prune the candidate space. 
The pruning is equivalent to a function that takes a KB $\mathcal{K}$ and a question $q$ as input,
and outputs a much limited set $\mathcal{C}$ of candidate subject-relation pairs. 
\begin{equation}\label{eq:pruning-function}
\mathcal{H}(\mathcal{K}, q) \to \mathcal{C}
\end{equation}
$\mathcal{C}_s$ and $\mathcal{C}_r$ are used to represent the subject and relation candidates, respectively.


The fundamental intuition for pruning is that the subject entity must be mentioned by some textual substring (\textit{subject mention}) in the question. 
Thus, the candidate space can be restricted to entities whose name/alias matches an n-gram of the question, as in~\cite{yih2014semantic,yih2015semantic,bordes2015large}.
We refer to this straight-forward method as \textit{N-Gram pruning}.
By considering all n-grams, this approach usually achieves a high recall rate. 
However, the candidate pool is still noisy due to many non-subject-mention n-grams. 

Our key idea is to reduce the noise by guiding the pruning method's attention to more probable parts of a question. 
An observation is that certain parts of a sentence are more likely to be the subject mention than others. 
For example, ``\textit{Harry Potter}'' in ``\textit{Who created the character Harry Potter}'' 
is more likely than ``\textit{the character}'', ``\textit{character Harry}'', etc.
Specifically, our method employs a deep network to identify such focus segments in a question. 
This way, the candidate pool can be not only more compact, but also significantly less noisy.

Finally, combing the ideas of Eq.\eqref{eq:conditional-view} and \eqref{eq:pruning-function}, we propose an approximate solution to the problem defined by Eq.~\eqref{eq:single-fact-qa-defn} 
\vspace{-0.5em}
\begin{equation}
\hat{s}, \hat{r} \approx \argmax_{s, r \in \mathcal{C}} p(s|q,r)p(r|q)
\label{eq:single-fact-approx-defn}
\end{equation}

%% file: 040model.tex
In this section, we first review the gated recurrent unit (GRU), an RNN variant extensively used in this work. 
Then, we describe the model parameterization of $p(r|q)$ and $p(s|q,r)$, and
the focused pruning method in inference. 

%% file: 041review_gru.tex

In this work we employ GRU~\cite{cho2014learning} as the RNN structure.
At time step $t$, a GRU computes its hidden state $h_t$ using the following compound functions
\begin{align}
  z &= \text{sigmoid}\left( W_{xz} x_t + W_{hz} h_{t-1} + b_z \right)\\ 
  r &= \text{sigmoid}\left( W_{xr} x_t + W_{hr} h_{t-1} + b_r \right)\\ 
  \tilde{h} &= \tanh \left( W_{xh} x_t + r \otimes W_{hh} h_{t-1} + b_h \right) \\
  h_t &= z \otimes h_{t-1} + (1-z) \otimes \tilde{h}
\end{align}
where $W_{\{\cdot\}}$, and $b_{\{\cdot\}}$ are all trainable parameters.
To better capture the context information on both sides, two GRUs with opposite directions can be combined to form a bidirectional GRU (BiGRU).


%% file: 042model_parameterization.tex
\paragraph{Relation network} 
In this work, the probability of relations given a question, $p(r|q)$, is modeled by the following network
\begin{equation}\label{eq:relation-probability}
p_{\theta_r}(r|q) = \frac{\exp\big(v(r,q)\big)}
{\sum_{r^\prime} \exp \big(v(r^\prime,q)\big)}
\end{equation}
where the \textit{relation scoring function} $v(r, q)$ measures the similarity between the question and the relation
\begin{equation}\label{eq:relation-score}
v(r,q) = f(q)^\top \embed{r}
\end{equation}
$\embed{r}$ is the trainable embedding of the relation (randomly initialized in this work) and $f(q)$ computes the semantic question embedding.
Specifically, the question $q$ is represented as a sequence of tokens (potentially with unknown ones).
Then, the question embedding model $f$ consists of 
a word embedding layer to transform tokens into distributed representations, 
a two-layer BiGRU to capture the question semantics, 
and a linear layer to project the final hidden states of the BiGRU into the same vector space as $\embed{r}$.


\paragraph{Subject network}
As introduced in section \ref{sec:overview}, the factor $p(s | q, r)$ models the fitness of a subject $s$ appearing in the question $q$, given the main topic is about the relation $r$.
Thus, two forces 
\begin{inparaenum}[\itshape a\upshape)]
	\item the raw context expressed by $q$, and
	\item the candidate topic described by $r$,
\end{inparaenum}
jointly impact the fitness of the subject $s$.
For simplicity, we use two additive terms to model the joint effect
\begin{equation}\label{eq:subject-probability}
p_{\theta{s}}(s|q,r) = \frac{\exp\big(u(s,r,q)\big)}
{\sum_{s^\prime} \exp\big(u(s^\prime,r,q)\big)}
\end{equation}
where $u(s,r,q)$ is the \textit{subject scoring function}, 
\begin{equation}\label{eq:subject-score}
u(s,r,q) = g(q)^\top \embed{s} + \alpha h(r,s) 
\end{equation}
$g(q)$ is another semantic question embedding, $\embed{s}$ is a vector representation of a subject, $h(r,s)$ is the subject-relation score, and $\alpha$ is the weight parameter used to trade off the two sources. 

Firstly, the \textit{context score} $g(q)^\top \embed{s}$ models the intrinsic plausibility that the subject $s$ appears in the question $q$ using vector space similarity.
As $g(q)^\top \embed{s}$ has the same form as equation \eqref{eq:relation-score}, 
we let $g$ adpot the same model structure as $f$.
However, initializing $\embed{s}$ randomly and training it with supervised signal, just like training $\embed{r}$, is insufficient in practice ---
while a large-scale KB has millions of subjects, only thousands of question-triple pairs are available for training.
To alleviate the problem, we seek two potential solutions: \textit{a) pretrained} embeddings, and \textit{b) type vector} representation. 


The pretrained embedding approach utilizes unsupervised method to train entity embedings. 
In particular, we employ the \texttt{TransE}~\cite{bordes2013translating}, which trains the embedings of entities and relations by enforcing $\embed{s} + \embed{r} = \embed{o}$ for every observed triple $(s,r,o) \in \mathcal{K}$.
As there exists other improved variants~\cite{gu2015traversing}, \texttt{TransE} scales the best when KB size grows.

Alternatively, type vector is a fixed (not trainable) vector representation of entities using type information. 
Since each entity in the KB has one or more predefined types, we can encode the entity as a vector (bag) of types.
Each dimension of a type vector is either 1 or 0, indicating whether the entity is associated with a specific type or not.
Thus, the dimensionality of a type vector is equal to the number of types in KB.
Under this setting, with $\embed{s}$ being a binary vector, let $g(q)$ be a continuous vector with arbitrary value range can be problematic. 
Therefore, when type vector is used as $\embed{s}$, we add a sigmoid layer upon the final linear projection of $g$, squashing each element of $g(q)$ to the range $[0, 1]$.

Compared to the first solution, type vector is fully based on the type profile of an entity, and requires no training.
As a benefit, considerably fewer parameters are needed.
Also, given the type information is discriminative enough, using type vector will lead to easier generalization.
However, containing only type information can be very restrictive.

\begin{figure}[tb]
	\centering
	\small
	\includegraphics[width=0.4\textwidth]{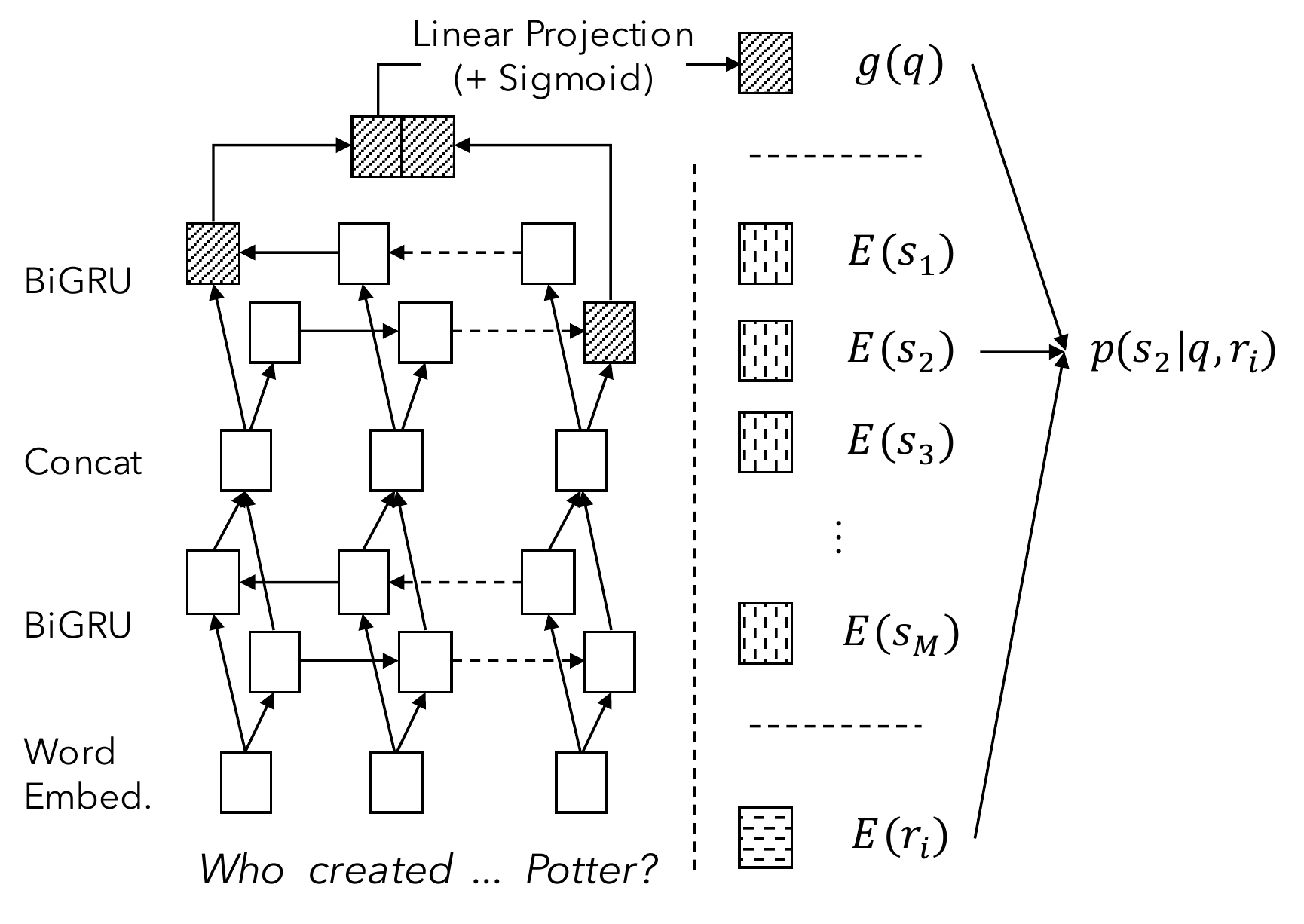}
	\caption{Overall structure of the subject network. Sigmoid layer is added only when type vector is used as $\embed{s}$.}
	\label{fig:subjct-score-function}
	\vspace{-0.5em}
\end{figure}

In addition to the context score, we use the \textit{subject-relation score} $h(r,s)$ to capture the compatibility that $s$ and $r$ show up together.
Intuitively, for an entity to appear in a topic characterized by a relation, a necessary condition will be that the entity has the relation connected to it. 
Inspired by this structural regularity, in the simplest manner, we instantiate the idea with an indicator function,
\begin{equation}
h(r,s) = \mathds{1}(s \to r)
\end{equation}
As there exists other more sophisticated statistical parameterizations, the proposed approach is able to capture the core idea of the structural regularity without any parameter. 
Finally, putting two scores together, Fig.\ref{fig:subjct-score-function} summarizes the overall structure of the subject network.

%% file: 043subject_labeling.tex
As discussed in section \ref{sec:pruning-formulation}, N-Gram pruning is still subject to large amount of noise in inference due to many non-subject-mention n-grams.
Motivated by this problem, we propose to reduce such noise by focusing on more probable candidates using a special-purpose sequence labeling network. 
Basically, a sequence labeling model is trained to tag some consecutive tokens as the subject mention.
Following this idea, during inference, only the most probable n-gram predicted by the model will be retained, and then used as the subject mention to generate the candidate pool $\mathcal{C}$.
Hence, we refer to this method as \textit{focused pruning}. 
Formally, let $\mathcal{W}(q)$ be all the n-grams of the question $q$, $p(w | q)$ be the probability that the n-gram $w$ is the subject mention of $q$, the focused pruning function $\mathcal{H}_s$ is defined as 
\begin{equation}
\begin{aligned}
\hat{w} &= \argmax_{w \in \mathcal{W}(q)} p_{\kappa}(w | q) \\
\mathcal{C} &= \{(s, r) : \mathcal{M}(s, \hat{w}), s \to r\}
\end{aligned}
\end{equation}
where $\mathcal{M}(s, \hat{w})$ represents some predefined match between the subject $s$ and the predicted subject mention $\hat{w}$.
Intuitively, this pruning method resembles the human behavior of first identifying the subject mention with the help of context, and then using it as the key word to search the KB.

To illustrate the effectiveness of this idea, we parameterize $p_{\kappa}(w|q)$ with a general-purpose neural labeling model, which consists of a word embedding layer, two layers of BiGRU, and a linear-chain conditional random field (CRF). 
Thus, given a question $q$ of length $T$, the score of a sequence label configuration $y \in \mathbb{R}^{T}$ is
\begin{equation*}
s(y, q) = \sum_{t=1}^{T} \mathbf{H}(q)_{t,y_t} +  \sum_{t=2}^{T} \mathbf{A}_{y_{t-1}, y_{t}}
\end{equation*}
where $\mathbf{H}(q)$ is the hidden output of the top-layer BiGRU, $\mathbf{A}$ is the transition matrix possesed by the CRF, and $[\cdot]_{i,j}$ indicates the matrix element on row $i$ collum $j$.

Finally, the match function $\mathcal{M}(s, \hat{w})$ is simply defined as either strict match between an alias of $s$ and $\hat{w}$, or approximate match provided by the Freebase entity suggest API~\footnote{The approximate match is used only when there is no strict match. The suggest API takes a string as input, and returns no more than 20 potentially matched entities.}. 
Note that more elaborative match function can further boost the performance, but we leave it for future work.

%% file: 050training.tex
In this section, we discuss the parameter estimation for the neural models presented in section \ref{sec:model}.

With standard parameterization, the focused labeling model $p_{\kappa}(w|q)$ can be directly trained by maximum likelihood estimation (MLE) and back-propagation.
Thus, we omit the discussion here, and refer readers to \cite{huang2015bidirectional} for details.
Also, we leave the problem of how to obtain the training data to section \ref{sec:experiments}.

%% file: 051decompose_likelihood.tex
To estimate the parameters of $p_{\theta_r}(r|q)$ and $p_{\theta_s}(s|r,q)$, MLE can be utilized to maximize the empirical (log-)likelihood of subject-relation pairs given the associated question. 
Following this idea, let $\{{s^{(i)}, r^{(i)}}, q^{(i)}\}_{i = 1}^N$ be the training dataset, the MLE solution takes the form
\begin{equation}\label{eq:mle-joint-solution}
\begin{aligned}
\theta^{\text{MLE}} =& \argmax_{\theta_r, \theta_s} \sum_{i=1}^{N} 
\Big( \log p_{\theta_r}(r^{(i)}|q^{(i)}) \\
&+ \log p_{\theta_s}(s^{(i)}|r^{(i)},q^{(i)}) \Big)
\end{aligned}
\end{equation}
Note that there is no shared parameter between $ p_{\theta_s}(s|q,r)$ and $p_{\theta_r}(r|q)$.~\footnote{Word embeddings are not shared across models.}
Therefore, the same solution can be reached by separately optimizing the two log terms, i.e.
\begin{eqnarray}\label{eq:mle-separate-solution}
\begin{aligned}
\theta_r^{\text{MLE}} &= \argmax_{\theta_r} \sum_{i=1}^{N} \log p_{\theta_r}(r^{(i)}|q^{(i)}) \\
\theta_s^{\text{MLE}} &= \argmax_{\theta_s} \sum_{i=1}^{N} \log p_{\theta_s}(s^{(i)}|r^{(i)},q^{(i)})
\end{aligned}
\end{eqnarray}
It is important to point out that the decomposability does not always hold. For example, when the parametric form of $h(s,r)$ depends on the embedding of $r$, the two terms will be coupled and joint optimization must be performed. 
From this perspective, the simple form of $h(s,r)$ also eases the training by inducing the decomposability.

%% file: 052hinge_loss.tex
As the two problems defined by equation \eqref{eq:mle-separate-solution} take the standard form of classification, theoretically, cross entropy can used as the training objective.
However, computing the partition function is often intractable, especially for $p_{\theta_s}(s|r,q)$, since there can be millions of entities in the KB.
Faced with this problem, classic solutions include contrastive estimation~\cite{smith2005contrastive}, importance sampling approximation~\cite{bengio2003neural}, and hinge loss with negative samples~\cite{collobert2008unified}.

In this work, we utilize the hinge loss with negative samples as the training objective.
Specifically, the loss function w.r.t $\theta_r$ has the form
\begin{equation}
\begin{aligned}
\mathcal{L}(\theta_r) = \sum_{i=1}^{N} \sum_{j=1}^{M_r}& \max\big[ 0, \gamma_r - v(r^{(i)}, q^{(i)}) \\
&+ v(r^{(j)}, q^{(i)})\big]
\end{aligned}
\end{equation}
where $r^{(j)}$ is one of the $M_r$ negative samples (i.e. $s^{(i)} \not\to r^{(j)}$) randomly sampled from $\mathcal{R}$, and $\gamma_r$ is the predefined margin.
Similarly, the loss function w.r.t $\theta_s$ takes the form
\begin{align}\label{eq:hinge-loss-sub}
\mathcal{L}(\theta_s) = \sum_{i=1}^{N} \sum_{j=1}^{M_s}& \max\big[ 0, \gamma_s - u(s^{(i)},r^{(i)},q^{(i)}) \notag \\
&+ u(s^{(j)},r^{(i)},q^{(i)})\big]
\end{align}
Despite the negative sample based approximation, there is another practical difficulty when type vector is used as the subject representation.
Specifically, computing the value of $u(s^{(j)},r^{(i)},q^{(i)})$ requires to query the KB for all types of each negative sample $s^{(j)}$.
So, when $M_s$ is large, the training can be extremely slow due to the limited bandwidth of KB query.
Consequently, under the setting of type vector, we instead resort to the following type-wise binary cross-entropy loss
\begin{equation}
\begin{aligned}
\tilde{\mathcal{L}}(\theta_s) = - \sum_{i=1}^{N} \sum_{k=1}^{K} \Big( \embed{s^{(i)}}_k \log g(q^{(i)})_k  \\
+ \big[1-\embed{s^{(i)}}_k\big] \log \big[1-g(q^{(i)})_k\big] \Big)
\end{aligned}
\end{equation}
where $K$ is the total number of types, $g(q)_k$ and $\embed{s^{(i)}}_k$ are the $k$-th element of $g(q)$ and $\embed{s^{(i)}}$ respectively.
Intuitively, with sigmoid squashed output, $g(q)$ can be regarded as $K$ binary classifiers, one for each type. 
Hence, $g(q)_k$ reprents the predicted probability that the subject is associated with the $k$-th type.

%% file: 070experiments.tex
In this section, we conduct experiments to evaluate the proposed system empirically.

%% file: 071dataset.tex
We train and evaluate our method on the  \textsc{SimpleQuestions} dataset\footnote{\url{https://research.facebook.com/researchers/1543934539189348}} --- 
the largest question-triple dataset. 
It consists of 108,442 questions written in English by human annotators. 
Each question is paired with a subject-relation-object triple from Freebase. 
We follow the same splitting for training (70\%), validation (10\%) and testing (20\%) as \cite{bordes2015large}.
We use the same subset of Freebase (\texttt{FB5M}) as our knowledge base so that the results are directly comparable. 
It includes 4,904,397 entities, 7,523 relations, and 22,441,880 facts. 

There are alternative datasets available, such as WebQuestions~\cite{berant2013semantic} and Free917~\cite{cai2013large}. However, these datasets are quite restricted in sample size --- 
the former includes 5,810 samples (train + test) and the latter includes 917 ones. 
They are fewer than the number of relations in Freebase.

%

To train the focused labeling model, the information about whether a word is part of the subject mention is needed. 
We obtain such information by reverse linking from the ground-truth subject to its mention in the question.
Given a question $q$ corresponding to subject $s$, we match the name and aliases of $s$ to all n-grams that can be generated from $q$.
Once a match is found, we label the matched n-gram as the subject mention. 
In the case of multiple matches, only the longest matched n-gram is used as the correct one.

%% file: 072evaluation.tex
For evaluation, we consider the same metric introduced in \cite{bordes2015large}, which takes the prediction as correct if both the subject and relation are correctly retrieved. 
Based on this metric, we compare \name with a few baseline systems, which include both the Memory Network QA system \cite{bordes2015large}, and systems with alternative components and parameterizations from existing work~\cite{yih2014semantic,yih2015semantic}. 
We did not compare with alternative subject networks because the only existing method~\cite{yih2014semantic} relies on unique textual name of each entity, 
which does not generally hold in knowledge bases (except in \textsc{Reverb}). 
Alternative approaches for pruning method, relation network, and entity representation are described below. 

\paragraph{Pruning methods} We consider two baseline methods previously used to prune the search space. 
The first baseline is the N-Gram pruning method introduced in Section~\ref{sec:overview}, as it has been successfully used in previous work~\cite{yih2014semantic,yih2015semantic}. 
Basically, it establishes the candidate pool by retaining subject-relation pairs whose subject can be linked to one of the n-grams generated from the question.
The second one is N-Gram+, a revised version of the N-Gram pruning with additional heuristics~\cite{bordes2015large}. 
Instead of considering all n-grams that can be linked to entities in KB, heuristics related to overlapping n-grams, stop words, interrogative pronouns, and so on are exploited to further shrink the n-gram pool. 
Accordingly, the search space is restricted to subject-relation pairs whose subject can be linked to one of the remaining n-grams after applying the heuristic filtering.


\paragraph{Relation scoring network}
We compare our proposed method with two previously used models. 
The first baseline is the embedding average model (Embed-AVG) used in \cite{bordes2014question,bordes2014open,bordes2015large}.
Basically, it takes the element-wise average of the word embeddings of the question to be the question representation.
The second one is the letter-tri-gram CNN (LTG-CNN) used in \cite{yih2014semantic,yih2015semantic}, where the question and relation are separately embedded into the vector space by two parameter shared LTG-CNNs.~\footnote{In Freebase, each predefined relation has a \textit{single} human-recognizable reference form, usually a sequence of words.}
In addition, \cite{yih2014semantic,yih2015semantic} observed better performance of the LTG-CNN when substituting the subject mention with a special symbol. Naturally, this can be combined with the proposed focused labeling, since the latter is able to identify the potential subject mention in the question. So, we train another LTG-CNN with symbolized questions, which is denoted as LTG-CNN+. Note that this model is only tested when the focused labeling pruning is used.


\paragraph{Entity representation} 
In section \ref{sec:model_parameterization}, we describe two possible ways to improve the vector representation of the subject, \texttt{TransE} pretrained embedding and type vectors. 
To evaluate their effectiveness, we also include this variation in the experiment, and compare their performance with randomly initialized entity embeddings.

%% file: 073experiment_setting.tex
During training, all word embeddings are initialized using the pretrained GloVe~\cite{pennington2014glove}, and then fine tuned in subsequent training. 
The word embedding dimension is set to 300, and the BiGRU hidden size 256. 
For pretraining the entity embeddings using \texttt{TransE} (see section \ref{sec:model_parameterization}), only triples included in \texttt{FB5M} are used.
All other parameters are randomly initialized uniformly from $[-0.08, 0.08]$, following \cite{graves2013generating}. 
Both hinge loss margins $\gamma_s$ and $\gamma_r$ are set to $0.1$. 
Negative sampling sizes $M_s$ and $M_r$ are both 1024.

For optimization, parameters are trained using mini-batch AdaGrad~\cite{duchi2011adaptive} with Momentum~\cite{pham2015optimization}.
Learning rates are tuned to be 0.001 for question embedding with type vector, 0.03 for LTG-CNN methods, and 0.02 for rest of the models.
Momentum rate is set to 0.9 for all models, and the mini-batch size is 256.
In addition, vertical dropout~\cite{pham2014dropout,zaremba2014recurrent} is used to regularize all BiGRUs in our experiment.~\footnote{For more details, source code is available at \url{http://zihangdai.github.io/cfo} for reference.}

%% file: 074results.tex
\begin{table}[tb]
	\small
	\centering	
	\tabcolsep=0.11cm
	\begin{tabular}{llccc}
		\toprule
		\multirow{2}{*}{\parbox{1cm}{\textbf{Pruning Method}}}
		& 
		\multirow{2}{*}{\parbox{1.5cm}{\textbf{Relation Network}}} &
		 \multicolumn{3}{c}{\textbf{Entity Representation}} \\
		\cmidrule(r){3-5}
		&
		& Random
		& Pretrain
		& Type Vec \\
		\midrule
		\multicolumn{2}{c}{Memory Network} & \multicolumn{3}{c}{62.9~~63.9$^*$} \\
		\midrule
		\multirow{3}{*}{N-Gram}
		& Embed-AVG        & 39.4 & 42.2 & 50.9 \\
		& LTG-CNN          & 32.8 & 36.8 & 45.6 \\
		& BiGRU            & 43.7 & 46.7 & 55.7 \\
		\midrule
		\multirow{3}{*}{N-Gram+}
		& Embed-AVG        & 53.8 & 57.0 & 58.7 \\
		& LTG-CNN          & 46.3 & 50.9 & 56.0 \\
		& BiGRU            & 58.3 & 61.6 & 62.6 \\
		\midrule
		\multirow{4}{*}{\parbox{1.01cm}{Focused Pruning}}
		& Embed-AVG        & 71.4 & 71.7 & 72.1 \\
		& LTG-CNN          & 67.6 & 67.9 & 68.6 \\
		& LTG-CNN+         & 70.2 & 70.4 & 71.1 \\
		& BiGRU            & 75.2 & 75.5 & \textbf{75.7} \\
		\bottomrule
	\end{tabular}
	\caption{Accuracy on \textsc{SimpleQuestions} testing set. 
	$^*$ indicates using ensembles. N-Gram+ uses additional heuristics. The proposed \name (focused pruning + BiGRU + type vector) achieves the top accuracy.}
	\label{tab:results}
\end{table}

\begin{table*}[htb]
	\centering
	\small
	\begin{tabular}{lcccc}
		\toprule
		\multirow{2}{*}{Pruning method}& \multirow{2}{1cm}{Pruning recall}& \multicolumn{2}{c}{Inference accuracy within the recalled} & \multirow{2}{1cm}{Overall accuracy} \\
		\cmidrule(r){3-4}
		 &  & Single-subject case & Multi-subject case & \\
		\midrule
		N-Gram              & 94.8\% & \quad 18 / 21 \quad\enspace = 85.7\% & 12051 / 20533 = 58.7\% & 55.7\% \\
		N-Gram+ & 92.9\% & \enspace 126 / 138 \quad = 91.3\%       & 13460 / 20017 = 67.2\% & 62.6\%  \\
		Focused pruning & 94.9\% & 9925 / 10705 = 92.7\%          & \, 6482  / 9876 \enspace = 65.6\%  & \textbf{75.7\%} \\		
		\bottomrule
\end{tabular}
	\caption{Comparison of different space pruning methods. N-Gram+ uses additional heuristics. Single- and multi-subject refers to the number of distinct subjects in candidates. The proposed focused pruning achieves best scores.}
	\label{tab:recall-precision}
\end{table*}

Trained on 75,910 questions, our proposed model and baseline methods are evaluated on the testing set with 21,687 questions. 
Table~\ref{tab:results} presents the accuracy of those methods. 
We evaluated all combinations of pruning methods, relation networks and entity representation schemes, as well as the result from memory network, as described in Section \ref{sec:dataset}.
\name (focused pruning + BiGRU + type vector) achieves the best performance, outperforming all other methods by substantial margins.

By inspecting vertically within each cell in Table~\ref{tab:results}, 
for the same pruning methods and entity representation scheme, BiGRU based relation scoring network boosts the accuracy by 3.5 \% to 4.8\% compared to the second best alternative.
This evidence suggests the superiority of RNN in capturing semantics of question utterances.
Surprisingly, it turns out that Embed-AVG achieves better performance than the more complex LTG-CNN.

By inspecting Table~\ref{tab:results} horizontally, 
type vector based representation constantly leads to better performance, especially when N-Gram pruning is used. 
It suggests that under sparse supervision, training high-quality distributed knowledge representations remains a challenging problem.
That said, pretraining entity embeddings with \texttt{TransE} indeed gives better performance compared to random initialization, indicating the future potential of unsupervised methods in improving continuous knowledge representation.

In addition, all systems using our proposed focused pruning method outperform their counterparts with alternative pruning methods. 
Without using ensembles, \name is already better than the memory network ensembles by 11.8\%. 
It substantiates the general effectiveness of the focused pruning with subject labeling method regardless of other sub-modules.


%% file: 075eval_subject_labeling.tex
According to the results in section \ref{sec:results}, the focused pruning plays a critical role in achieving the best performance. 
To get a deeper understanding of its effectiveness, we analyze how the pruning methods affect the accuracy of the system.
Due to space limit, we focus on systems with BiGRU as the relation scoring function and type vector as the entity representation.

Table \ref{tab:recall-precision} summarizes the recall --- the percentage of pruned subject-relation candidates containing the answer --- and the resulting accuracy. 
The single-subject case refers to the scenario that there is only one candidate entity in $\mathcal{C}_{s}$ (possibly with multiple relations), and the multi-subject case means there are multiple entities in $\mathcal{C}_{s}$.
As the table shows, focused pruning achieves comparable recall rate to N-Gram pruning.\footnote{Less than 3\% of the recalled candidates rely on approximate matching in the focused pruning.}
Given the state-of-the-art performance of sequence labeling systems, this result should not be surprising.
Thus, the difference in performances entirely comes from their resulting accuracy.
Notice that there exists a huge accuracy gap between the two cases.
Essentially, in the single-candidate case, the system only need to identify the relation based on the more robust model $p_{\theta_r}(r | q)$. 
In contrast, under the multi-candidate case, the system also relies on $p_{\theta_s}(s | q, r)$, which has significantly more parameters to estimate, and thus is less robust.
Consequently, by only focusing on the most probable sub-string, the proposed focused pruning produces much more single-candidate situations, leading to a better overall accuracy.

%% file: 076additional_analysis.tex
In the aforementioned experiments, we have kept the focused labeling model and the subject scoring network fixed.
To further understand the importance and sensitivity of this specific model design, we investigate some variants of these two models.

\paragraph{Alternative focus with CRF}
RNN-CRF based models have achieved the state-of-the-art performance on various sequence labeling tasks~\cite{huang2015bidirectional,lu2015twisted}.
However, the labeling task we consider here is relatively unsophisticated in the sense that there are only two categories of labels - part of subject string (\texttt{SUB}) or not (\texttt{O}).
Thus, it's worth investigating whether RNN (BiGRU in our case) is still a critical component when the task gets simple.
Hence, we establish a CRF baseline which uses traditional features as input.
Specifically, the model is trained with \texttt{Stanford CRF-NER} toolkit~\footnote{\url{http://nlp.stanford.edu/software/CRF-NER.shtml}} on the same reversely linked labeling data (section \ref{sec:dataset}).
For evaluation, we directly compare the sentence level accuracy of these two models on the test portion of the labeling data.
A sentence labeling is considered correct only when all tokens are correctly labeled.~\footnote{As $F$-1 score is usually used as the metric for sequence labeling, sentence level accuracy is more informative here.}
It turns out the RNN-CRF achieves an accuracy of 95.5\% while the accuracy of feature based CRF is only 91.2\%.
Based on the result, we conclude that BiGRU plays a crucial role in our focused pruning module.

\paragraph{Subject scoring with average embedding}
As discussed in section \ref{sec:model_parameterization}, the subject network $g$ is chosen to be the same as $f$, mainly relying on a two-layer BiGRU to produce the semantic question embeding.
Although it is a natural choice, it remains unclear whether the final performance is sensitive to this design.
Motivated by this question, we substitute the BiGRU with an Embed-AVG model, and evalute the system performance.
For this experiment, we always use focused pruning and type vector, but vary the structure of the relation scoring network to allow high-order interaction across models.
The result is summarized in Table \ref{tab:subject_scoring_network_analysis}.
Insepcting the table horizontally, when BiGRU is employed as the subject network, the accuracy is consistently higher regardless of relation network structures.
However, the margin is quite narrow, especially compared to the effect of varying the relation network structure the same way.
We suspect this difference reflects the fact that modeling $p(s | r, q)$ is intrinsically more challenging than modeling $p(r|q)$.
It also suggests that learning smooth entity representations with good discriminative power remains an open problem.

\begin{table}[tb]
	\small
	\centering	
	\tabcolsep=0.11cm
	\begin{tabular}{lcc}
		\toprule
		\multirow{2}{*}{\parbox{3cm}{\textbf{Relation Network}}} &
		\multicolumn{2}{c}{\textbf{Subject Network}} \\
		\cmidrule(r){2-3}
		& Embed-AVG
		& BiGRU \\
		\midrule
		Embed-AVG      & 71.6 & 72.1 \\
		LTG-CNN          & 68.0 & 68.6 \\
		LTG-CNN+        & 70.4 & 71.1 \\
		BiGRU               & 75.4 & 75.7 \\
		\bottomrule
	\end{tabular}
	\caption{System performance with different subject network structures.}
	\label{tab:subject_scoring_network_analysis}
\end{table}

%% file: 080conclusion.tex
In this paper, we propose \name, a novel approach to single-fact question answering. 
We employ a conditional factoid factorization by inferring the target relation first and then the target subject associated with the candidate relations. 
To resolve the representation for millions of entities, we proposed type-vector scheme which requires no training. 
Our focused pruning largely reduces the candidate space without loss of recall rate, leading to significant improvement of overall accuracy. 
Compared with multiple baselines across three aspects, our method achieves the state-of-the-art accuracy on a 108k question dataset, the largest publicly available one.
Future work could be extending the proposed method to handle more complex questions. 